\documentclass[11pt,a4paper]{article}
\usepackage[hyperref]{acl2019}
\usepackage{times}
\usepackage{latexsym}
\usepackage{subfigure}
\usepackage{url}
\usepackage{bm}
\usepackage{natbib}
\usepackage{graphicx}
\usepackage{multirow}
\usepackage{booktabs}
\usepackage{stfloats}
\usepackage{verbatim}
\usepackage{CJKutf8}

\aclfinalcopy % Uncomment this line for the final submission
\usepackage{footmisc}
\title{Automatic Generation of Personalized Comment Based on User Profile}
\author{Wenhuan Zeng\textsuperscript{1}\thanks{\ \ Equal Contribution.}, 
Abulikemu Abuduweili\textsuperscript{2}\footnotemark[1], 
Lei Li\textsuperscript{3}, 
Pengcheng Yang\textsuperscript{4}\\
\textsuperscript{1}School of Mathematical Sciences, Peking University\\
\textsuperscript{2}State Key Lab of Advanced Optical Communication System and Networks,\\ School of EECS, Peking University\\
\textsuperscript{3}School of Computer Science and Technology, Xidian University \\
\textsuperscript{4}MOE Key Lab of Computational Linguistics, School of EECS, Peking University\\
  \texttt{\{zengwenhuan, abduwali\}@pku.edu.cn} \\
  \texttt{tobiaslee@foxmail.com, yang\_pc@pku.edu.cn} \\
  }

\begin{document}
\begin{CJK}{UTF8}{gbsn}
\maketitle
\begin{abstract}
%With the blooming of the Internet, social media has gradually become a significant communication tool in the recent years.  Automatically generating comments in a personalized way is necessary for promoting the user experience on social media. However, 
Comments on social media are very diverse, in terms of content, style and vocabulary, which make generating comments much more challenging than other existing natural language generation~(NLG) tasks. Besides, since different user has different expression habits, it is necessary to take the user's profile into consideration when generating comments.  
In this paper, we introduce the task of automatic generation of personalized comment~(AGPC) for social media. Based on tens of thousands of users' real comments and corresponding user profiles on weibo, we propose Personalized Comment Generation Network~(PCGN) for AGPC. The model utilizes user feature embedding with a gated memory and attends to user description to model personality of users. In addition, external user representation is taken into consideration during the decoding to enhance the comments generation. Experimental results show that our model can generate natural, human-like and personalized comments.\footnote{Source codes of this paper are available at \href{https://github.com/Walleclipse/AGPC}{https://github.com/Walleclipse/AGPC}}
% The codes can be found at 
\end{abstract}

\section{Introduction}
Nowadays, social media is gradually becoming a mainstream communication tool. People tend to share their ideas with others by commenting, reposting or clicking \textit{like} on posts in social media. Among these behaviors, \textit{comment} plays a significant role in the communication between posters and readers. Automatically generate personalized comments~(AGPC) can be useful due to the following reasons. First, AGPC helps readers express their ideas more easily, thus make them engage more actively in the platform. Second, bloggers can capture different attitudes to the event from multiple users with diverse backgrounds. Lastly, the platform can also benefit from the increasing interactive rate.

% 讨论理论的可行性
Despite its great applicability, the AGPC task faces two important problems: whether can we achieve it and how to implement it? 
The \textit{Social Differentiation Theory} proposed by \newcite{Riley} proved the feasibility of building a universal model to automatically generate personalized comments based on part of users’ data. 
The \textit{Individual Differences Theory} pointed by \newcite{Hovland} answers the second question by introducing the significance of users' background, which inspires us to incorporating user profile into comments generation process. More specifically, the user profile consists of demographic features~(for example, where does the user live), individual description and the common word dictionary extracted from user's comment history. 
% 这个必要性可以不提了吧？
%综上，有结合个人档案，针对社交媒体自动生成具有特定用户个人色彩的评论有进行研究的必要。
There are few works exploring the comments generation problem. \citet{Zheng} first paid attention to generating comments for news articles by proposing a gated attention neural network model~(GANN) to address the contextual relevance and the diversity of comments. Similarly, \citet{Qin} introduced the task of automatic news article commenting and released a large scale Chinese corpus. Nevertheless, AGPC is a more challenging task, since it not only requires generating relevant comments given the blog text, but also needs the consideration of the diverse users' background. 

% there is little work on generating relevant comments of data collected from social media such as Weibo, where the personalities of users are playing a significant role.
%本篇论文中我们提出了一个自动生成个性化评论的创新任务。任务的目标是结合有用户人口统计学因素作为用户的个人档案对社交媒体上的内容生成带有用户个人色彩的评论，对于有历史评论的用户，提取其常用词生成常用词词典，作为其个人档案的补充。同时我们介绍了一个大规模数据集，原始数据集包含有个用户对条博文生成的条评论。我们贡献的数据集与现有数据集不同的地方在于一下几点：
  In this paper, we propose a novel task, automatically generating personalized comment based on user profile.
  We build the bridge between user profiles and social media comments based on a large-scale and high-quality Chinese dataset. We elaborately design a generative model based on sequence-to-sequence~(Seq2Seq) framework. A gated memory module is utilized to model the user personality. Besides, during the decoding process, the model attends to user description to enhance the comments generation process.
  In addition, the vocabulary distribution of generated word is adapted by considering the external user representation.

%介绍提出的方法（abduwali）
% To express personality naturally and coherently in a comments, we design a sequence to sequence generation model equipped with new mechanisms. That mechanisms includes gated memory for leveraging grammatical correctness and personalized style,  blog-user co-attention for efficiently exploit user individual descriptions while decoding, and an external personality expression by concatenation of user description and decoding output to help generate more explicit personalized commenting style.
% With the provided dataset and the seq2seq framework with above mechanisms, we believe that this work can advance the development of NLG tasks based on non-canonical languages and promote the research based on language influenced social and demographic factors.

%小结（第二，三条 abduwali）
Our main contributions are as follows:
\begin{itemize}
\item We propose the task of automatic generating personalized comment with exploiting user profile.
\item We design a novel model to incorporate the personalized style in large-scale comment generation. The model has three novel mechanisms: user feature embedding with gated memory, blog-user co-attention, and an external personality expression.
\item Experimental results show that the proposed method outperforms various competitive baselines by a large margin. With novel mechanisms to exploit user information, the generated comments are more diverse and informative. We believe that our work can benefit future work on developing personalized and human-like NLG model.
\end{itemize}

\section{Personalized Comments Dataset}
\begin{figure}[t]
    \centering
    \includegraphics[width=1.0\linewidth]{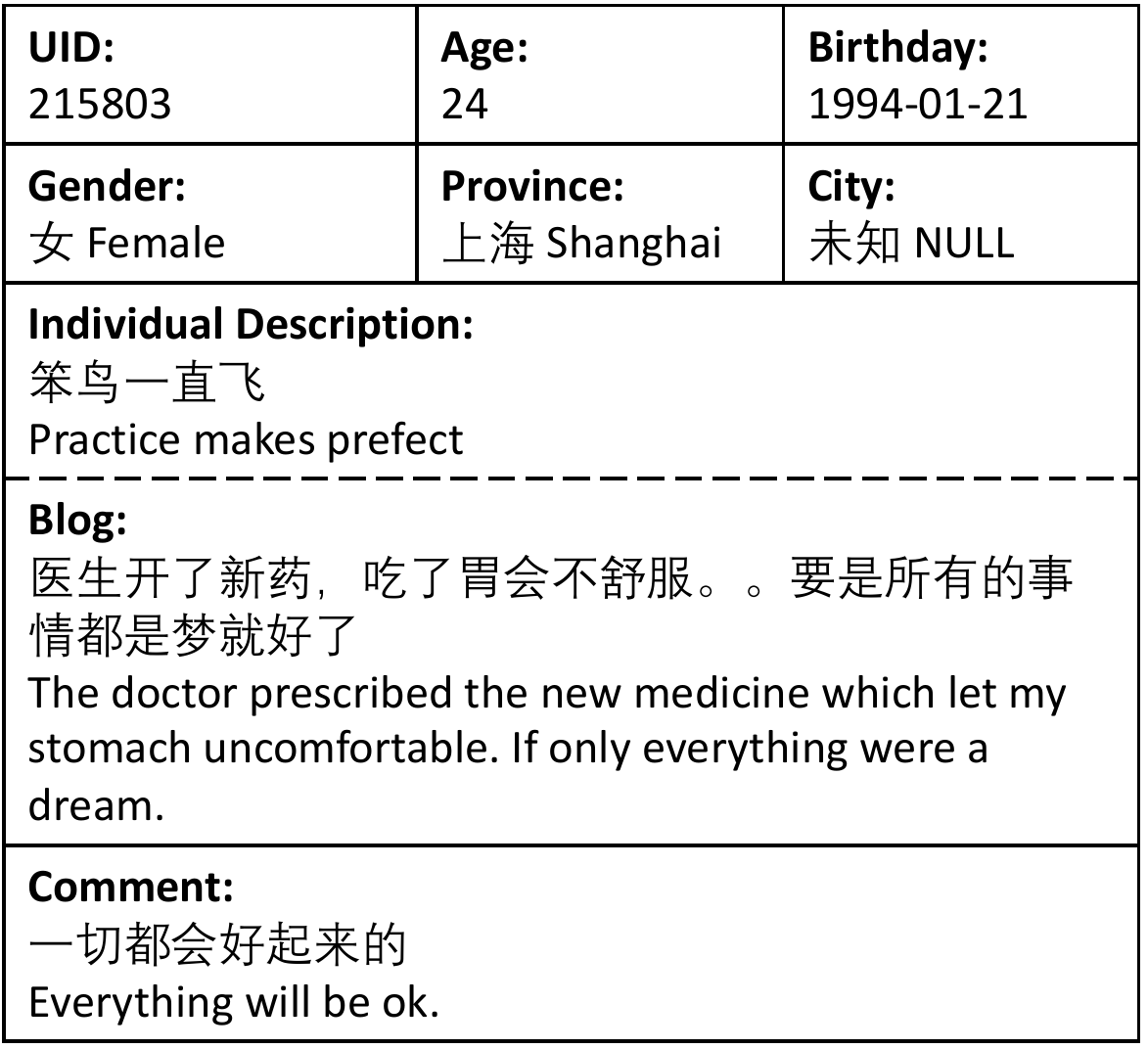}
    \caption{A data example in personalized comment dataset. Corresponding English translation is provided.}
    \label{fig:example}
    \vspace{-0.15in}
\end{figure}

%在这一部分，我们将从以下几方面对个性化评论数据集做介绍。
We introduce the dataset as follows:
\paragraph{Data Preparation} We collect short text posts from Weibo, one of the most popular social media platform in China, which has hundreds of millions of active users.
%Table 1 展示了我们获取到的数据中的一条样本实例，为便于阅读，我们给出了对应的英文翻译，由Table1可知，我们的每一条实例包含了用户的省市，性别，年龄，婚姻状况，个人描述，用户发表的评论及对应博文。
Each instance in the dataset has province, city, gender, age, marital status, individual description of user's, comment added by user and homologous blog content. Figure~\ref{fig:example} visually shows a sample instance.
%本文使用jieba对每条样本的评论，博文和个人描述进行了分词，将文本变成单词。为了方便模型从数据集中学习到有效信息，我们去除了评论，博文及个人描述中的@标识，网址，表情并将中文统一成简体字
 We tokenized all text like individual description, comment and blog content into words, using a popular python library Jieba\footnote{\href{https://github.com/fxsjy/jieba}{https://github.com/fxsjy/jieba}}. To facilitate the model to learn valid information from the dataset, we removed @, url, expressions in the text, and unified Chinese into simplified characters.
%对省份、城市、性别和婚姻状况这样的离散变量采用one-hot编码的方式统一处理。
 Discrete variables such as province, city, gender and marital status were treated uniformly by one-hot coding.
%经过数据清洗后，从控制样本质量的角度考虑，我们保留评论及博文所含单词的个数均为两个及以上的样本。为了从用户的人口统计学特征中学习到用户独特的表达习惯，数据集中每个用户均有50条及以上的记录。最后，我们得到的数据集涵盖了238,854 位用户对 3215088 条博文发表条的 11042084 评论。
  To ensure the quality of text, we filtered out samples with less than two words in the variable of comment and blog content. Besides, in order to learn  user-specific expression habits, we retain users with 50 or more records. The resulting dataset contains 4,463,767 comments on 1,495,822 blog posts by 32,719 users. 
\paragraph{Data Statistics} We split the corpus into training, validation and testing set according to the microblog. To avoid overfitting, the records of the same microblog will not appear in the above three sets simultaneously.
  %我们按照博文将数据集划分为训练集，验证集及测试集三部分。为了避免过拟合的情况出现，对同一个博文的评论不会同时出现在训练集、验证集和测试集中。
  %Table2展示了训练集、验证集何测试集含个人档案数、评论数及博文数的情况。
  Table~\ref{tab:statstics_of_dataset} displays the detail sample size of user, comment and blog about training set, validation set and testing set.
%经过对我们得到的数据集中文本型变量的统计，每个用户平均有46条样本。博文平均含49.65个单词，评论平均含11.28个单词，个人描述平均含8.85个单词,对各个实验数据集的详细统计如Table3表所示
  Each user in the resulting dataset has an average of 56 samples. The average lengths of blog post, comment and individual description are 50, 11 and 9 words, respectively. The particular statistics of each experimental dataset are shown in Table~\ref{tab:text_variables_statistics}.
%实际场景中，并不是所有用户都会完整的填写涉及到隐私的个人信息，以致于我们获取到的数据集的用户信息部分存在缺失值，为了实验数据的真实性和丰富性，本文仅对生日变量中的缺失值进行均值插补，保留其余变量中的缺失值。关于数据集缺失值的统计结果展示在表4中。
% Blog 的英文也要加上去
%%%%%%%%%%%%%%%%%% dataset  sample %%%%%%%%%%%%%%%%
% \begin{table*}[t]
% \begin{tabular}{llllllll}
% \toprule
% Uid & Age & Birthday & Gender & Marital & Province & City & Individual Description~~~~~~~~~ \\ 
% \midrule
% 215803 & 20 & 1998-03-26 & \begin{tabular}[c]{@{}l@{}}男性\\ Male\end{tabular} & \begin{tabular}[c]{@{}l@{}}单身\\ Single\end{tabular} & \begin{tabular}[c]{@{}l@{}}山东\\ Shandong\end{tabular} & \begin{tabular}[c]{@{}l@{}}淄博\\ Zibo\end{tabular} & 帅气的情感忍者 \\ 
% \end{tabular}
% \begin{tabular}{ll}
% \midrule
% Blog & Comment \\ 
% \midrule
% 不要太在意别人对自己的看法，有些人有嘴，但不一定有脑！ & 说的对，要活出自己的风采 \\ 
% \bottomrule
% \end{tabular}
% \caption{A data example in personalized comment dataset. Corresponding English translation is provided.}
% \label{tab:example_of_dataset}
% \vspace{-0.05in}
% \end{table*}
%%%%%%%%%%%%%%%%%%%% dataset sample %%%%%%%%%%%%%%%%%%%%%
%Notably, there is some missing values in user profile, since users would not fill their privacy-related individual information completely. However, considering the authenticity and richness of the experimental data, we only performs the mean interpolation on the missing values in the 'birthday' entry, and retains the missing values in the remaining variables. 
%The ratios of missing values for each variables are shown in Table~\ref{tab:missing_value_statistics}.
%%%%%%%%%%%%%%%%%%%%%%%% dataset %%%%%%%%%%%%%%%%%%%%%%%%%%%%
\begin{table}[t]
\begin{center}
\begin{tabular}{|l|c c c|}
\hline \textbf{Statistic} & \textbf{User} & \textbf{Comment} & \textbf{Microblog}\\ 
\hline
\ \textbf{Train} & 32,719 & 2,659,870 & 1,450,948\\
\ \textbf{Dev} & 24,739 & 69,659 & 27,822\\
\ \textbf{Test} & 20,157 & 43,866 & 17,052\\
\ \textbf{Total} & 32,719 & 4,463,767 & 1,495,822\\
\hline
\end{tabular}
\end{center}
\caption{\label{tab:statstics_of_dataset} Sample size of three datasets }
\vspace{-0.15in}
\end{table}
%%%%%%%%%%%%%%%%%%%%%%%% dataset %%%%%%%%%%%%%%%%%%%%%%%%%%%%
\begin{table}[h]
\begin{center}
\begin{tabular}{|l|c c c c|}
\hline \textbf{Average length} & \textbf{Train} & \textbf{Dev} & \textbf{Test} & \textbf{Total}\\ 
\hline
\ ID & 8.84 & 9.04 & 8.83 & 8.85 \\
\ Comment & 11.28 & 11.32 & 11.86 & 11.28 \\
\ Microblog & 49.67 & 47.95 & 50.30 & 49.65 \\
\hline
\end{tabular}
\end{center}
\caption{\label{tab:text_variables_statistics} Statistics of text variables. Individual description, abbreviated ID.}
\vspace{-0.15in}
\end{table}
%%%%%%%%%%%%% dataset %%%%%%%%%%%%%%%%%%%%%%%%%%%%%%%%%%%%%%%
%\begin{table}[t]
%\begin{center}
%\begin{tabular}{|l|c c c c|}
%\hline \textbf{Statistic} & \textbf{Train} & \textbf{Dev} & %\textbf{Test} & \textbf{Total}\\  \hline
%\ Province & 0.0021 & 0.0020 & 0.0020 & 0.0021 \\
%\ City & 0.0021 & 0.0020 & 0.0020  & 0.0021 \\
%\ Gender & 0.0091 & 0.0083 & 0.0097 & 0.0091 \\
%\ MS & 0.0116 & 0.0114 & 0.0110 & 0.0116 \\
%\ ID & 0.0857 & 0.0864 & 0.0793 & 0.0857 \\
%\hline
%\end{tabular}
%\end{center}
%\caption{\label{tab:missing_value_statistics} Missing value ratio of the user profile. MS denotes marital status and ID means description. }
%\vspace{-0.2in}
%\end{table}
%\paragraph{Quality of comments}\quad\\

\section{Personalized Comment Generation Network}
\begin{figure*}[t]
    \vspace{-10pt}
    \centering  %插入的图片居中表示	
    \includegraphics[width=0.95\linewidth]{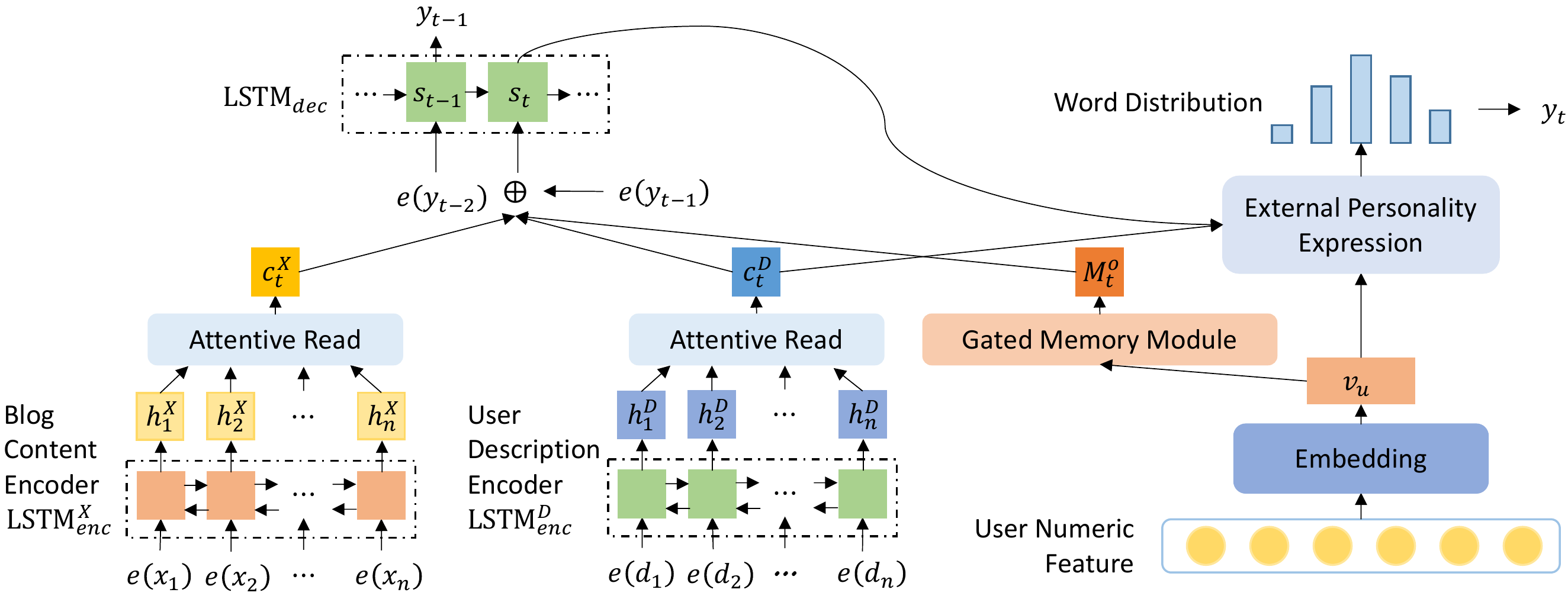}
    \caption{\label{fig:PCGN} Personalized comment generation network }
    \vspace{-12pt}
\end{figure*}

% \subsection{Problem Definition}
Given a blog $X=(x_1,x_2,\cdots, x_n)$ and a user profile $U=\{F,D\}$, where $F=(f_1,f_2,\cdots,f_k)$ denotes the user's numeric feature~(for example, age, city, gender) and $D=(d_1,d_2,\cdots,d_l)$ denotes the user's individual description, the AGPC aims at generating comment $Y=(y_1,y_2,\cdots,y_m)$ that is coherent with blog $X$ and user $U$. 
Figure \ref{fig:PCGN} presents an overview of our proposed model, which is elaborated on in detail as follows.
% In short, the model estimates the probability: $P(Y|X,U) = \prod_t P(y_t|y_{<t},X,U)$
% Building upon the general
% Based on encoder-decoder framework , we propose the Personalized Comment Generation Network (PCGN) to generate personalized comments using three mechanisms: First, we assume that during decoding, there is an internal user-profile state, and in order to capture the implicit change of the state and to balance the weights between the grammar state and the user-profile state dynamically, PCGN adopts an gated memory module to embed users. Second, some user's individual description is highly correlated to which commenting style. During decoding, in addition to context attention, we propose user description attention. Similarly, if common words for user commenting are given, we can also adopt attention mechanism to common words. Thirdly, after decoding, we concatenate the decoding output with user representation to explicit expression of an personality. An overview of PCGN is given in Figure \ref{fig:PCGN}.

\subsection{Encoder-Decoder Framework}
Our model is based on the encoder-decoder framework of the general sequence-to-sequence (Seq2Seq) model \cite{sutskever2014sequence}.
% It is implemented
% with bi-directional Long-Short Term Memory (LSTM)\cite{hochreiter1997long}. 
The encoder converts the blog sequence $X=(x_1, x_2, \cdots , x_n)$ to hidden representations $h^X =
(h^X_1, h^X_2, \cdots , h^X_n)$ by a bi-directional Long Short-Term Memory~(LSTM) cell \cite{hochreiter1997long}:
\begin{equation}
h^X_t = {\rm LSTM_{enc}^X}(h^X_{t-1}, x_t)
\end{equation}
%我觉得这里可以区分测试集和训练集，如果是训练集的话t时刻decoder的隐藏层st接受的是目标句子t-1时刻的单词，ct和s(t-1)。测试集接受的是上一时刻被预测出的y(t-1),ct和s(t-1).--wenhuan
% 不用考虑的这么细致 -- lilei
The decoder takes the embedding of a previously decoded word $e(y_{t-1})$ and a blog context vector $c^X_t$ as input to update its state $s_t$:
\begin{equation}
s_t = {\rm LSTM_{dec}}(s_{t-1},[c^X_t;e(y_{t-1})]) \label{eq:lstm_dec}
\end{equation}
where $[\cdot;\cdot]$ denotes vector concatenation.
% , serving as the input to the LSTM cell. 
The context vector $c^X_t$ is a weighted sum of encoder's hidden states, which carries key information of the input post~\cite{bahdanau2014neural}. 
Finally, the decoder samples a word $y_{t}$ from the output probability distribution as follows
% Once the state vector $s_t$ is obtained, the decoder
% generates a token by sampling from the output probability distribution $o_t$ computed from the decoder’s state st as follows:
%此处是tanh(st)还是st?--wenhuan
\begin{eqnarray}
y_t \sim {\rm softmax}(\mathbf{W_o} s_t)
% o_t &=& P(y_t | y_1,y_2,\cdots,y_{t-1},c_t) \\
%             &=& {\rm softmax}(W_o s_t)
\label{eq:dec_out}            
\end{eqnarray}
where \bm{$W_o$} is a weight matrix to be learned. 
The model is trained via maximizing the log-likelihood of ground-truth $Y^\ast = (y_1^\ast, \cdots, y_n^\ast)$ and the objective function is defined as 
\begin{eqnarray}
\mathcal{L} = - \sum_{t=1}^n{\rm log}\Big(p(y_t^*|y_{<t}^\ast, X, U)\Big)
\label{eq:loss}            
\end{eqnarray}

\subsection{User Feature Embedding with Gated Memory}
%user profile 不包含个人描述的意思？（abduwali）
% Method: Static user feature embedding -> Gated Memory 
% 简化说明 围绕方法的核心： embedding + dynamically expression
To encode the information in user profile, we map user's numeric feature $F$ to a dense vector $v_u$ through a fully-connected layer. Intuitively, $v_u$ can be treated as a user feature embedding denotes the character of the user.
% The most intuitive approach to modeling personality in comment generation is to take as additional input the user profile of a comment to be generated. Each user's embedding $v_u$ is represented by a real-valued, low dimensional vector. For user's numeric feature $F$, we learn the vectors of the user feature embedding through training (via fully connected layer). 
% But the method of user feature embedding is rather static: 
% the user feature embedding will not change during the generation process which may sacrifice grammatical correctness of sentences as argued in \cite{ghosh2017affect}. 
However, if the user feature embedding is static during decoding, the grammatical correctness of sentences generated may be sacrificed as argued in \citet{ghosh2017affect}.
To tackle this problem, we design an gated memory module to dynamically express personality during decoding, inspired by \citet{zhou2018emotional}.
Specifically, we maintain a internal personality state during the generation process. 
At each time step, the personality state decays by a certain amount. 
Once the decoding process is completed, the personality state is supposed to decay to zero, which indicates that the personality is completely expressed.
% We simulate the process of expressing personality as follows: there is an internal personality state for each user before the decoding process starts; 
% at each step the personality state decays by a certain amount; once the decoding process is completed, the personality state should decay to zero indicating the personality is completely expressed.
Formally, at each time step $t$, the model computes an update gate $g^u{_t}$ according to the current state of the decoder $s_{t}$.
The initial personality state $M_0$ is set as user feature embedding $v_u$. Hence, the personality state $M_t$ is erased by a certain amount (by $g_t^u$ ) at each step. This process is described as
\begin{eqnarray}
g^u_t &=& {\rm sigmoid}(\mathbf{W_g^u} s_t) \\
M_0 &=& v_u \\
M_t &=& g^u_t \otimes  M_{t-1} , \quad  t>0
\end{eqnarray}
where $ \otimes$ denotes element-wise multiplication. 
% 解释一下为什么要再加一个门
% 每一步都可以选择是否要表达 personality
Besides, the model should decide how much attention should be paid to the personality state at each time step. Thus, output gate $g_t^o$ is introduced to control the information flow by considering the previous decoder state $s_{t-1}$, previous target word $e(y_{t-1})$ and the current context vector $c^X_t$
\begin{equation}
     g^o_t  = {\rm sigmoid}(\mathbf{W_g^o} [s_{t-1};e(y_{t-1});c^X_t]).
\end{equation}
By an element-wise multiplication of $g_t^o$ and $M_t$, we can obtain adequate personality information $M_t^o$ for current decoding step
\begin{equation}
     M^o_t = g^o_t \otimes  M_t.
\end{equation}
% 这段不能少
%Finally, the state update mechanism in Eq.  \ref{eq:lstm_dec} is modified to
%\begin{equation}
%    s_t = {\rm LSTM_{dec}}(s_{t-1},[c_t;e(y_{t-1});M^o_t]).
%\end{equation}
% During decoding, PCGN also computes the output gate $g{_t}^o$ of memory module with the input of the word embedding of the previously decoded word $e(y_{t-1})$, the previous state of the decoder $s_{t-1}$, and the current context vector $c_t$. 
% LSTM updates its state $s_t$ conditioned on the previous target word $e(y_{t-1})$, the previous state of the decoder $s_{t-1}$, the context vector $c_t$, and the gated personality state output $M^o_t$,
% \begin{eqnarray}
% && g^o_t  = {\rm sigmoid}(W_g^o [s_{t-1};e(y_{t-1});c_t]) \\
% && M^o_t = g^o_t \otimes  M_t \\
% && s_t = {\rm LSTM_{dec}}(s_{t-1},[c_t;e(y_{t-1});M^o_t])
% \end{eqnarray}
% Based on the state, the word generation distribution $o_t$ can be obtained with Equation \ref{eq:dec_out} , and the next word $y_t$ can be sampled. After generating the next word, memory $M_t$ is updated.
\subsection{Blog-User Co-Attention}
Individual description is another important information source when generating personalized comments. 
% 这里的例子一定要具体到人吗 可以不可以用一般的明星代替
For example, a user with individual description ``只爱朱一龙''~(I only love Yilong Zhu\footnote{A famous Chinese star.}), tends to writes a positive and adoring comments on the microblog related to Zhu. 
% 这个名字 Blog -user 但是说明的顺序却是 user description 然后 blog
Motivated by this, we propose Blog-user co-attention to model the interactions between user description and blog content.
More specifically, we encode the user's individual description $D=(d_1, d_2, \cdots , d_l)$ to hidden states $(h^D_1, h^D_2, \cdots , h^D_l)$ via another LSTM
\begin{equation}
h^D_t = {\rm LSTM^D_{enc}}(h_{t-1}^D, d_t)
\end{equation}
We can obtain a description context vector $c_t^D$ by attentively reading the hidden states of user description, 
\begin{eqnarray}
c^D_t &=& \sum_j \alpha_{tj} h^D_j \\
\alpha_{tj} &=& {\rm softmax}(e_{tj}) \\
e_{tj} &=& s_{t-1} \mathbf{W_a}  h^D_j 
\end{eqnarray}
where $e_{tj}$ is a alignment score~\cite{bahdanau2014neural}.
% 这个 c_t^X 其实就是原来的 c_t 考虑要不要把 Seq2seq 中的 c_t 去掉 然后在这里再说这一点
Similarly, we can get the blog content vector $c_t^X$. Finally, the context vector $c_t$ is a concatenation of $c_t^X$ and $c_t^D$, in order provide more comprehensive information of user's personality
\begin{equation}
c_t = [c^X_t;c^D_t]
\end{equation}
Therefore, the state update mechanism in Eq.(\ref{eq:lstm_dec}) is modified to
\begin{equation}
     s_t = {\rm LSTM_{dec}}(s_{t-1},[c_t;e(y_{t-1});M^o_t])
\end{equation}

\subsection{External Personality Expression}
In the gated memory module, the correlation between the change of the internal personality state and selection of a word is implicit. 
% 这句话和这里的关联性不大
% Different users have different commenting styles and preferences for different words.  
To fully exploit the user information when selecting words for generation, we first compute a user representation $r_t^u$ with user feature embedding and user description context. 
\begin{equation}
r^u_t = \mathbf{W_r}[v_u;c^D_t]
\end{equation}
where $\mathbf{W_r}$ is a weight matrix to align user representation dimention.

The final word is then sampled from output distribution based on the concatenation of decoder state $s_t$ and $r_t^u$ as
\begin{equation}
\tilde{y_t} \sim \rm{softmax} ( \mathbf{W_{\tilde{o_t}}} [s_t; r_t^u] )
% \tilde{y_t} \sim  {\rm softmax } (\bf{W_{\tilde{o_t}}} {[s_t; r_t^u]} )
\end{equation}
where $\mathbf{W_{\tilde{o_t}}}$ is a learnable weight matrix.
% We propose an external personality expression module to more fully embody the personalized comment style. At each step $t$, PCGN compute a user representation with user feature embedding and user description context. 
% \begin{equation}
% r^u_t = [v_u;c^D_t]
% \end{equation}
% After decoding, we concatenate the decoding output $s_t$ with user representation to explicit expression of an personality.
% \begin{equation}
% \tilde{s_t} = [s_t;r^u_t]
% \end{equation}
% Using this refined  state vector $\tilde{s_t}$, with the mapping of fully connected layer, a token by sampling from the output probability distribution $\tilde{o_t}$ computed as:
% \begin{equation}
% \tilde{y_t} \sim \tilde{o_t} = {\rm softmax }(\bf{W_{\tilde{o_t}}} \tilde{s_t})
% \end{equation}

\section{Experiments}
\subsection{Implementation}
% Decoder 确定是双向的?
% We used Tensorflow\footnote{https://github.com/tensorflow/tensorflow} to implement our model. 
The blog content encoder and comment decoder are both 2-layer bi-LSTM with 512 hidden units for each layer.
% and use different sets of parameters respectively. 
The user's personality description encoder is a single layer bi-LSTM with 200 hidden units. 
The word embedding size is set to 300 and vocabulary size is set to 40,000. 
The embedding size of user's numeric feature is set to 100. 

We adopted beam search and set beam size to 10 to promote diversity of generated comments. 
We used SGD optimizer with batch size set to 128 and the learning rate is 0.001.

To further enrich the information provided by user description, we collected most common $k$  words in user historical comments~($k=20$ in our experiment). We concatenate the common words with the user individual description. Therefore, we can obtain more information about users' expression style. The model using concatenated user description is named PCGN with common words (PCGN+ComWord).
% We used the stochastic gradient descent~(SGD) algorithm with mini-batch. Batch size and learning rate are set to 128 and 0.001, respectively.
% to form a new user description. The model with concatened
\subsection{Baseline}
% As aforementioned, this paper is the first work to address the personalized comment generation in large-scale data set. We did not find closely-related baselines in the literature. Nevertheless, we chose two suitable baselines: 
We implemented a general Seq2Seq model \cite{sutskever2014sequence} and a user embedding model~(Seq2Seq+Emb) proposed by \citet{li2016persona} as our baselines. 
The latter model embeds user numeric features into a dense vector and feeds it as extra input into decoder at every time step.
% created by us where the user numeric features is embedded into a vector, and the vector serves as an input to every decoding position, similar to the idea of user embedding in\cite{li2016persona}, 
% Which is a proper baseline for our model.

\subsection{Evaluation Result}
\quad \textbf{Metrics:} 
We use BLEU-2~\cite{papineni2002bleu} and METEOR~\cite{banerjee2005meteor} to evaluate overlap between outputs and references. Besides, perplexity is also provided.
% Since our dataset is personalized comment dataset, both content level (whether the content is relevant and grammatical) and personality level (whether the content is relevant the commenting style of a given user) can be evaluated by general NLG metrics. The more accurate the grammar and the more personalized the commenting style, the similar the output hypothesis is to the correct reference. We adopted perplexity, BLEU2\cite{papineni2002bleu} and METEOR\cite{banerjee2005meteor} to evaluate our model.
% 这应该算是 Ablation Study 的结果
% 去掉 mem module 去掉 co-attn 结果 去掉 external 结果
% \textbf{Models:} In addition to baseline, We design four models, that apply different mechanisms respectively. 1. S2S + Mem: apply gated memory to embed user numeric features based one seq2seq framework. 2. S2S + Mem + CoAtt: compared with the previous model, apply Blog-User Co-Attention during decoding.  3. PCGN: compared with the previous model, apply external personality expression, which is concatenate the decoder output with user representation. 4. PCGN + ComWord: We collected common $k$(e.g. $k=20$ in our experiment) words in user historical comments. We concatenate the common words with user individual description and replace the individual description in PCGN with this new concatenated description. Other details is similar with previous PCGN model.

\begin{table}[t]
\begin{tabular}{llll}
\toprule
\textbf{Method}        & \textbf{PPL} & \textbf{B-2} & \textbf{METEOR} \\ \midrule
Seq2Seq       & 32.47      & 0.071   & 0.070  \\
Seq2Seq+Emb       & 31.13      & 0.084   & 0.079  \\ 
% \midrule
% S2S+Mem       & 30.73      & 0.099   & 0.083  \\
% S2S+Mem+CoAtt & 27.12      & 0.147   & 0.128  \\
\midrule
PCGN        & 27.94      & 0.162   & 0.132  \\ 
PCGN+ComWord & \textbf{24.48}      & \textbf{0.193}   & \textbf{0.151} \\ 
\bottomrule
\end{tabular}
\caption{Automatic evaluation results of different methods. \textbf{PPL} denotes perplexity and \textbf{B-2} denotes BLEU-2. Best results are shown in bold.}
\label{tab:eval_result}
\vspace{-0.15in}
\end{table}

\begin{table}[t]
\begin{tabular}{lll}
\toprule
\textbf{Method}        & \textbf{PPL} & \textbf{B-2} \\  %& \\ \textbf{METEOR} \\ 
\midrule
Seq2Seq       & 32.47       & 0.071  \\ %& 0.070  \\
~+~Mem       & 30.73 (-1.74)  & 0.099 (+0.028)  \\ % & 0.083  \\
~+~CoAtt    & 27.12 (-3.61)    & 0.147 (+0.078) \\%  & 0.128  \\
~+~External  & 27.94 (+0.82)  & 0.162 (+0.015)  \\%& 0.132  \\ 
%~+~ComWord  & 24.48 (-3.46)   & 0.193 (+0.031)  \\%& 0.151  %\\ 
\bottomrule
\end{tabular}
\caption{Incremental experiment results of proposed model. Performance on METEOR is similar to B-2. \textbf{Mem} denotes gated memory, \textbf{CoAtt} denotes blog-user co-attention and \textbf{External} denotes external personality expression}
\label{tab:incre_result}
\vspace{-0.15in}
\end{table}

\textbf{Results:} The results are shown in Table \ref{tab:eval_result}. As can be seen, PCGN model with common words obtains the best performance on perplexity, BLEU-2 and METEOR. Note that the performance of Seq2Seq is extremely low, since the user profile is not taken into consideration during the generation, resulting repetitive responses. In contrast, with the help of three proposed mechanism~(gated memory, blog-user co-attention and external personality expression), our model can utilize user information effectively, thus is capable of generating diverse and relevant comments for the same blog.
%\footnote{We present some generated personalized comments in appendix.}
Further, we conducted incremental experiments to study the effect of proposed mechanisms by adding them incrementally, as shown in Table~\ref{tab:incre_result}. 
It can be found that all three mechanism help generate more diverse comments, while blog-user co-attention mechanism contributes most improvements. An interesting finding is that external personality expression mechanism causes the decay on perplexity. We speculate that the modification on word distribution by personality influence the fluency of generated comments.

\section{Related Work}
%  Our approach builds on the previous work of following fields. 
%\paragraph{machine commenting}\quad\\
%本文的研究内容隶属于自然语言生成中的评论生成，先前研究中也有不少对评论生成任务的尝试。评论生成任务又可分为对结构数据生成评论、对文本数据生成评论、对图像生成评论以及对视频生成评论。
This paper focuses on comments generation task,
% which is a sub-task of natural language generation~(NLG) problem. 
% The comment generation task 
which can be further divided into generating a comment according to the structure data~\cite{Mei}, text data~\cite{Qin}, image~\cite{Vinyals} and video~\cite{ma2018livebot}, separately. 

%   The research content of this paper belongs to the generation of comments in NLG. 
%   The comment generation task can be further divided into generating a comment according to the structure data~\cite{Mei}, text data~\cite{Qin}, image~\cite{Vinyals} and video~\cite{ma2018livebot}, separately. 
% 工作焦点的区别 
There are many works exploring the problem of text-based comment generation. \citet{Qin} contributed a high-quality corpus for article comment generation problem. \citet{Zheng} proposed a gated attention neural network model~(GANN) to generate comments for news article, which addressed the contextual relevance and the diversity of comments. 
To alleviate the dependence on large parallel corpus, \citet{ma2018topic} designed an unsupervised neural topic model based on retrieval technique. 
However, these works focus on generating comments on news text, while comments on social media are much more diverse and personal-specific. 
% 篇幅不够了 这个社交媒体的研究不是特别相关
% Although some research have been conducted to analyze user behaviour on social media~\cite{Guo2009,guan2014analyzing,xue2014study}, there is little attention has been paid to generating comments on social media.
% 具体技术的区别

In terms of the technique for modeling user character, the existing works on machine commenting only utilized part of users' information. \citet{Ni} proposed to learn a latent representation of users by utilizing history information. \citet{Lin} acquired readers' general attitude to event mentioned by article through its upvote count. Compared to the indirection information obtained from history or indicator, 
user features in user profile, like demographic factors, can provide more comprehensive and specific information, and thus should be paid more attention to when generating comments.
Sharing the same idea that user personality counts, \citet{luo2018learning} proposed personalized MemN2N to explore personalized goal-oriented dialog systems. Equipped with a profile model to learn user representation and a preference
model learning user preferences, the model is capable of generating high quality responses. In this paper, we focus on modeling personality in a different scenario, where the generated comments is supposed to be general and diverse.
% In detail, the model consits 
% leverages user profiles and preferences to generate high quality response. 
% The PROFILE MODEL learns user personalities with
% distributed profile representation,
% with a similar idea to us.
% Sharing the same idea with us.
%   However, in the dialogue task, there existing studies have similar starting points with this paper.
%luo
% ~\citeauthor{luo2018learning} proposed a novel model named PERSONALIZED MEMN2N which leverage user profiles and preferences to make high quality goal-oriented dialogue come true. The model consists of a profile model and a preference model. Eventually, it's capable choose a proper language style and change recommendation policy based on the user profile and address the problem of ambiguity, separately.
%   While there have been a few attempts to generate nature language in a personalized way, to the best of our knowledge, no such prior work has been attempted for generating comment based on social media. We will draw on the ideas of our predecessors to complete our innovative task.
\section{Conclusion}
In this paper, we introduce the task of automatic generating personalized comment. 
We also propose Personality Comment Generation Network~(PCGN) to model the personality influence in comment generation.
The PCGN model utilized gated memory for user feature embedding, blog-user co-attention, and external personality representation to generate comments in personalized style. 
Evaluation results show that PCGN outpeforms baseline models by a large margin. 
With the help of three proposed mechanisms, the generated comments are more fluent and diverse.
% can generate relevant and personalized comments.
% responses not only in content (grammatically) but also in personalized commenting style.
\bibliography{acl2019}
\bibliographystyle{acl_natbib}

\appendix
\section{Case Study}
\begin{figure*}[h!]
    \centering
    \includegraphics[width=1.0\linewidth]{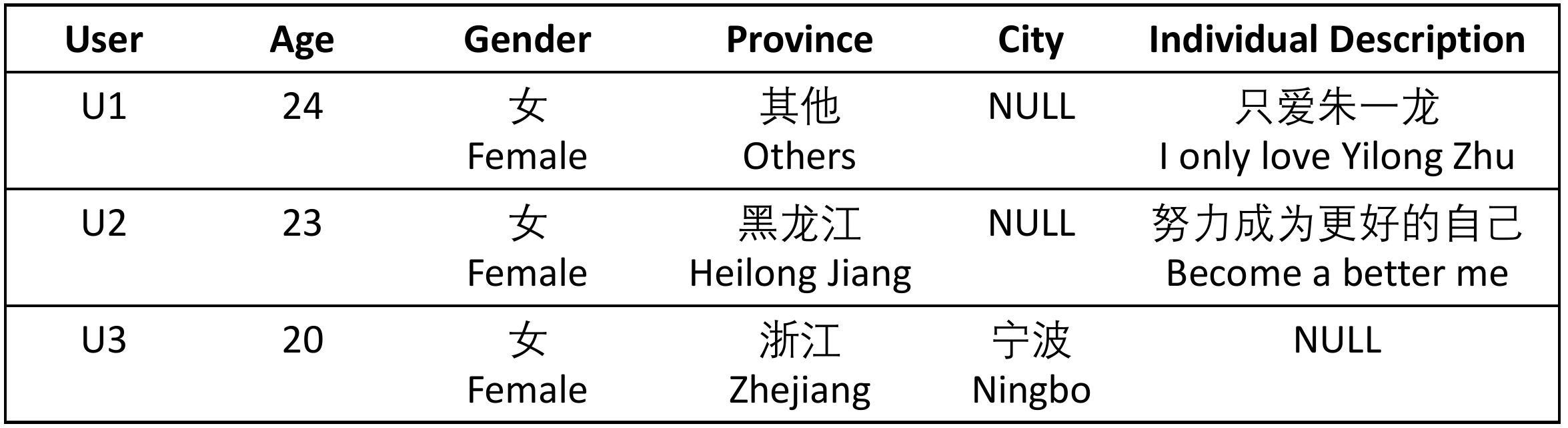}
    \caption{Part of user profile of case study users. In order to protect user privacy, the birthday variable is not shown here.}
    \label{fig:user}
\end{figure*}

\begin{figure*}[h!]
    \centering
    \includegraphics[width=1.0\linewidth]{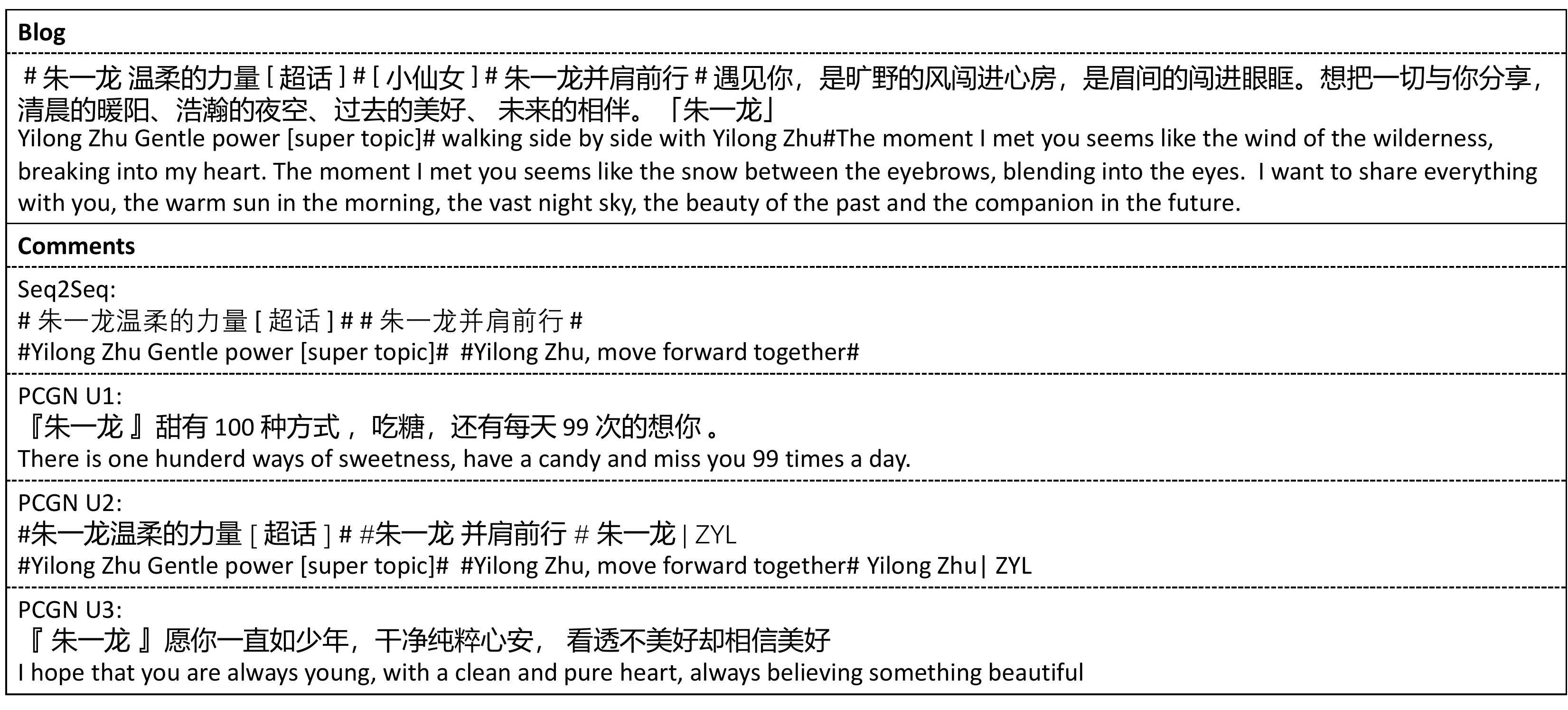}
    \caption{Generated comments based on blog of different users. Since Seq2Seq model does not take user profile into consideration, it generates same comments for the same blog.}
    \label{fig:case1}
    \vspace{-0.1in}
\end{figure*}

\begin{figure*}[tbh!]
    \centering
    \includegraphics[width=1.0\linewidth]{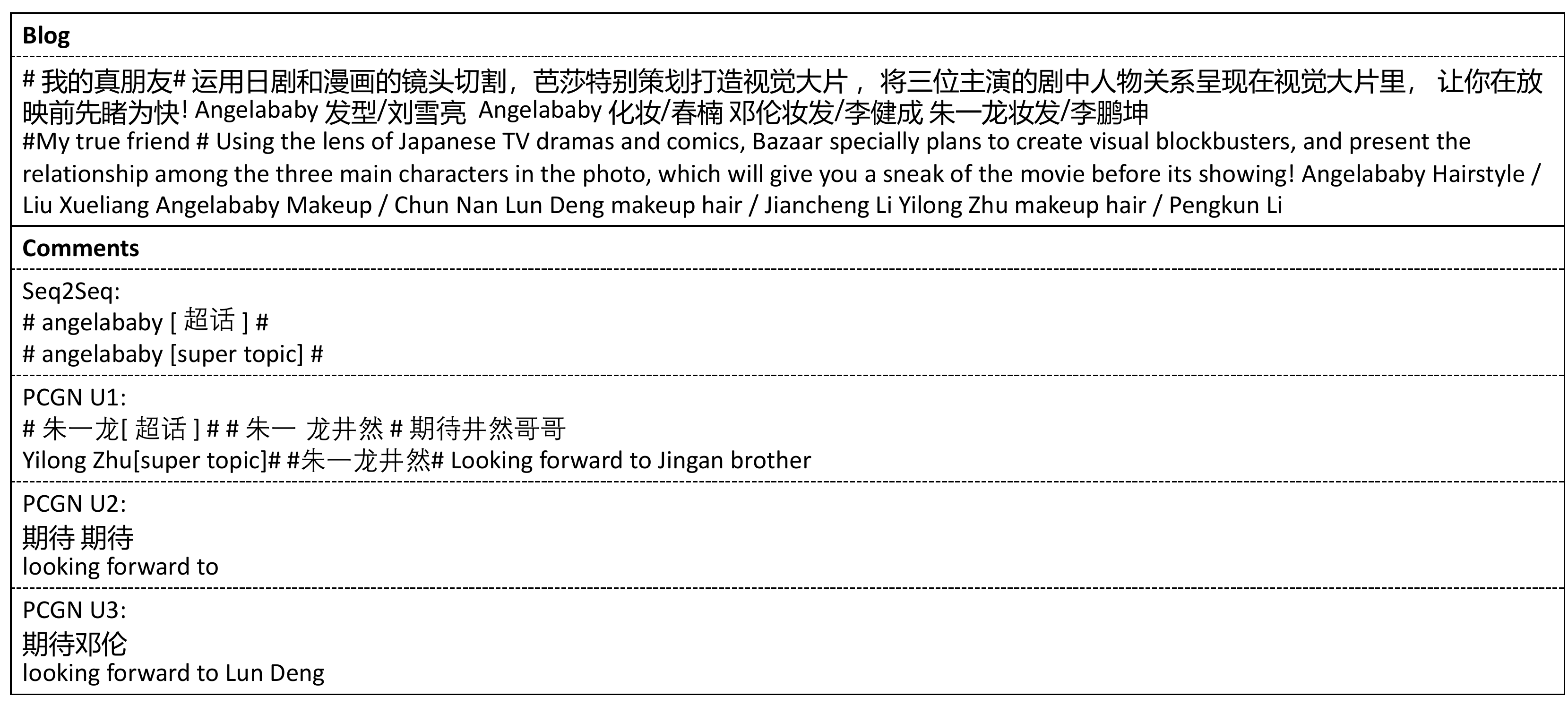}
    \caption{Generated comments based on blog of different users.}
    \label{fig:case2}
\end{figure*}

We present some generated cases in Figure \ref{fig:case1}, \ref{fig:case2}. There are multiple users (corresponding profiles are shown in Figure \ref{fig:user}) that are suitable for generating comments. Seq2Seq generates same comments for the same blog, while PCGN can generate personalized comment conditioned on given user. According to the user profile, U1 adores Yilong Zhu very much. 
% So for the blogs related to Yilong Zhu, 
Therefore, U1 tends to express her affection in comments when responses to blogs related to Yilong Zhu. For users whose individual descriptions can not offer helpful information or there is missing value for individual description, the PCGN model pays more attention to numeric features and learns representation from similar seen users.  


\end{CJK}
\end{document}

\begin{CJK}{UTF8}{gbsn}
\maketitle
\section{Case Study}
\begin{figure*}[h!]
    \centering
    \includegraphics[width=1.0\linewidth]{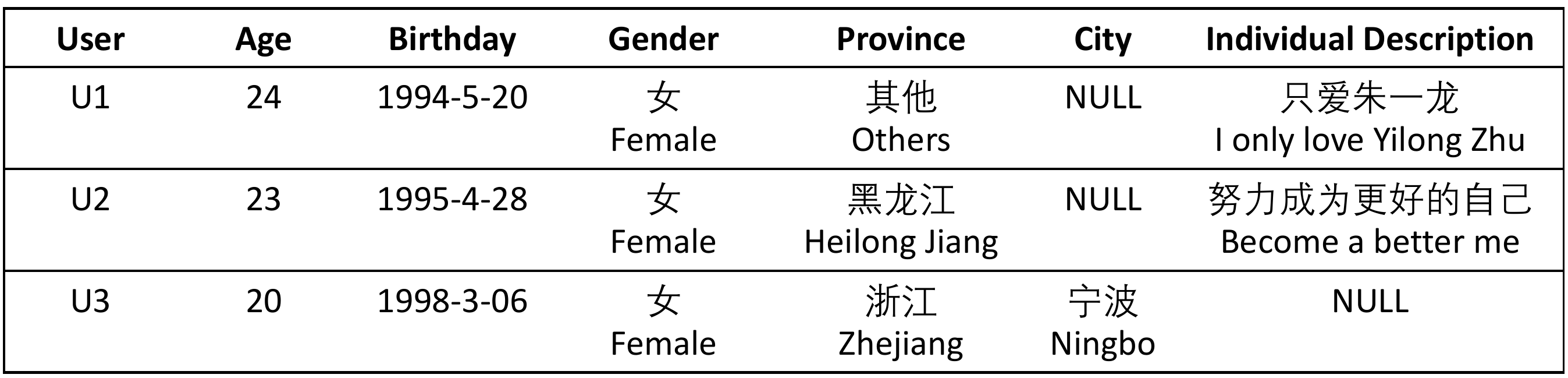}
    \caption{Part of user profile of case study users.}
    \label{fig:user}
\end{figure*}

\begin{figure*}[h!]
    \centering
    \includegraphics[width=1.0\linewidth]{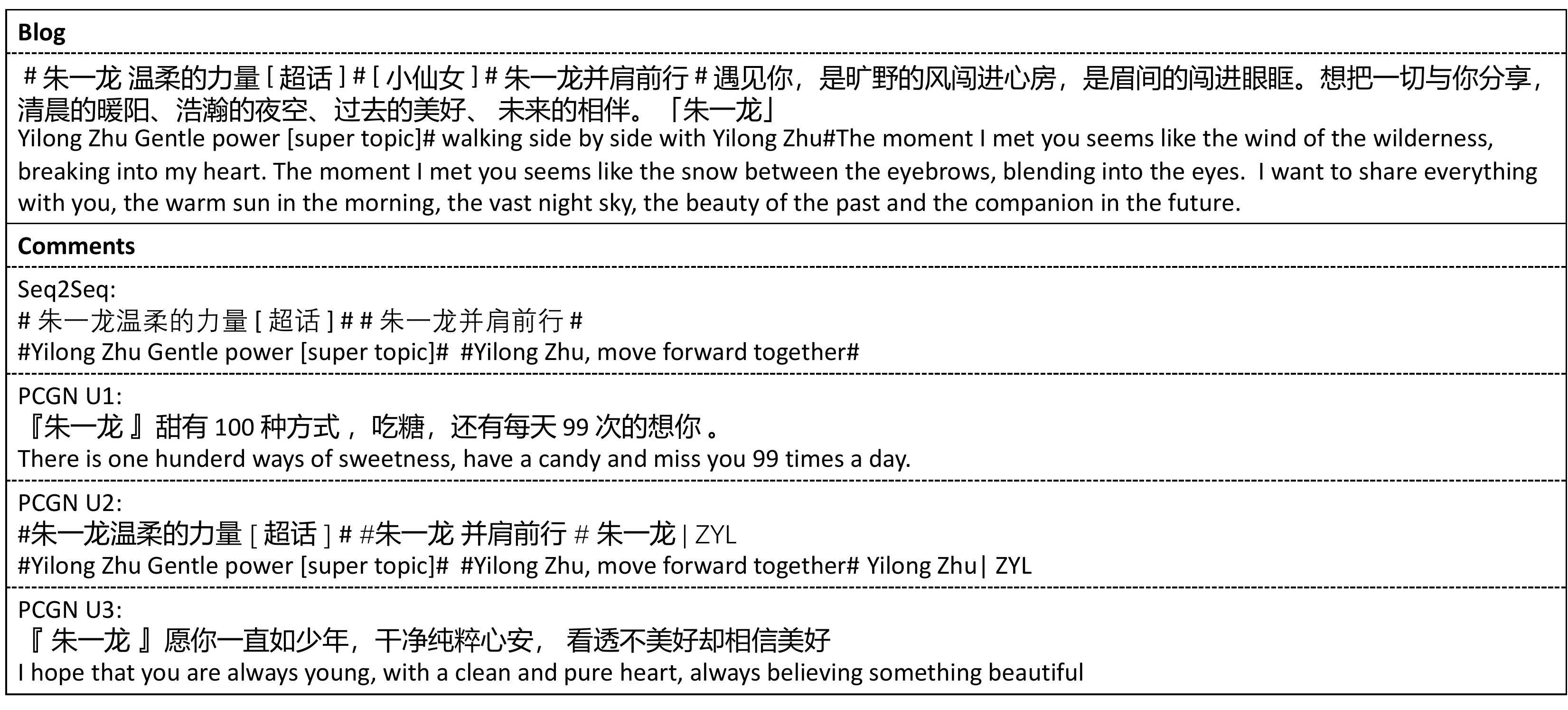}
    \caption{Generated comments based on blog of different users. Since Seq2Seq model does not take user profile into consideration, it generates same comments for the same blog.}
    \label{fig:case1}
\end{figure*}

\begin{figure*}[tbh!]
    \centering
    \includegraphics[width=1.0\linewidth]{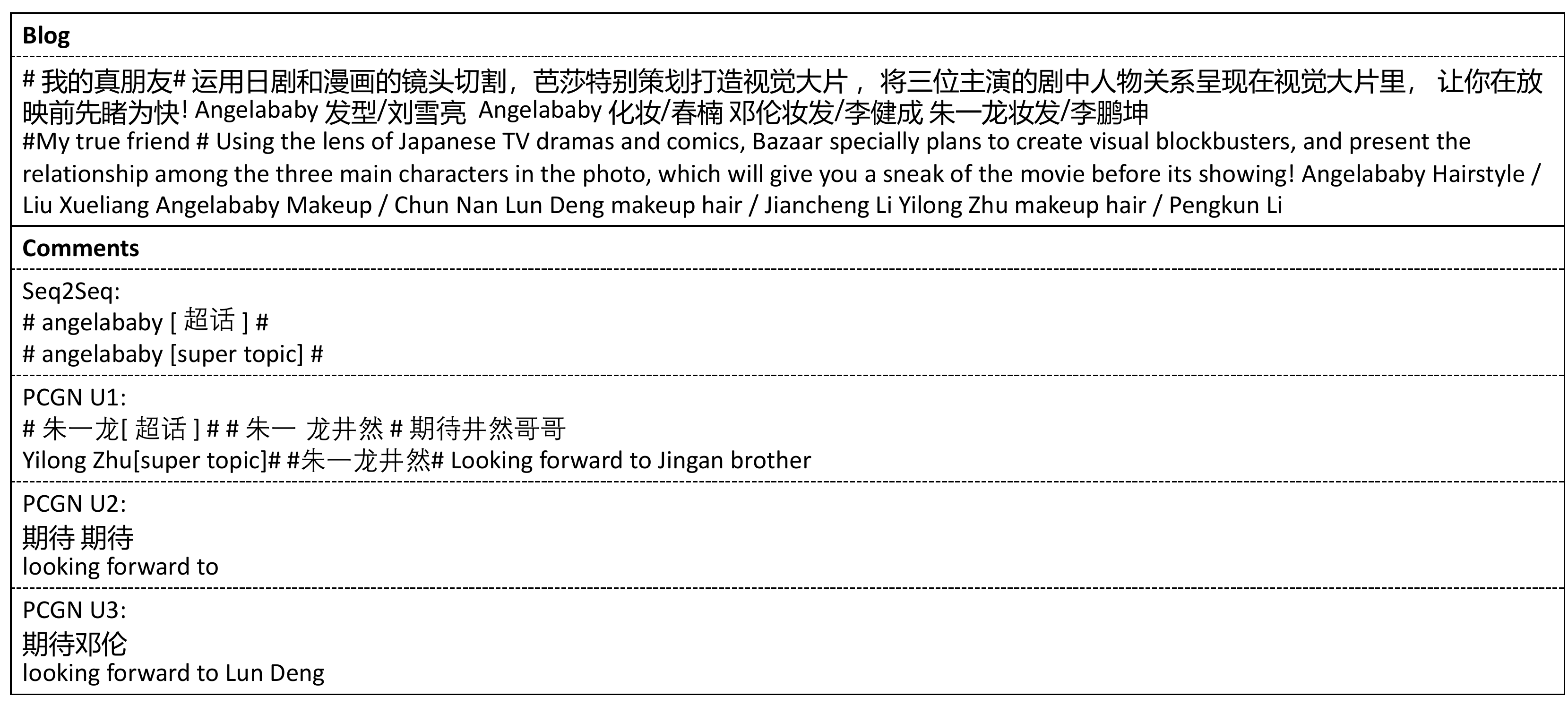}
    \caption{Generated comments based on blog of different users.}
    \label{fig:case2}
\end{figure*}

We present some examples in Figure \ref{fig:case1}, \ref{fig:case2}. As can be seen, for a given blog, there are multiple users (users profile are shown in Figure \ref{fig:user}) that are suitable for its response in comments. Seq2Seq generates a comment with a random user. PCGN can generate personalized comment conditioned on given user. According to the user profile, user 1 is more adores Yilong Zhu\footnote{A famous Chinese star}, So for the blogs related to Yilong Zhu , user 1 tend to express their affection in comments. For some users, their individual description can not offer helpful information or there is missing value for individual description. In this case, the PCGN model pay more attention to numeric features and learned representation from similar seen users.  

\newpage

% \begin{table*}[]
% \begin{tabular}{l|l|l|l|l}
%  \hline
%  \multirow{2}{*}{Blog} & \multirow{2}{*}{Seq2Seq} & \multicolumn{3}{l|}{PCGN} \\ \cline{3-5} 
%   &  & User 1 & User 2 & User 3 \\ \hline
% \#朱一龙温柔的力量{[}超话{]}\# {[}小仙女{]} \#朱一龙并肩前行\# 遇见你，是旷野的风闯进心房，是眉间的雪融进眼眶。想把一切与你分享，清晨的暖阳、浩瀚的夜空、过去的美好、未来的相伴 & \#朱一龙温柔的力量{[}超话{]}\# \#朱一龙并肩前行\# & 『朱一龙』甜有100种方式，吃糖，还有每天99次的想你。 & \#朱一龙温柔的力量{[}超话{]}\#\#朱一龙并肩前行\#朱一龙|JYL & 『朱一龙』愿你一直如少年，干净纯粹心安，看透不美好却相信美好 \\ \hline
%  \#我的真朋友\# 运用日剧和漫画的镜头切割， 芭莎特别策划打造视觉大片， 将三位主演的剧中人物关系呈现在视觉大片里， 让你在放映前先睹为快! Angelababy发型/刘雪亮  Angelababy化妆/春楠 邓伦妆发/李健成 朱一龙妆发/李鹏坤 & \#angelababy{[}超话{]}\# & \#朱一龙{[}超话{]}\#\#朱一龙井然\#期待井然哥哥 & 期待 期待 & 期待 邓伦 \\ \hline
%  \end{tabular}
%  \caption{Sample Comments generated by PCGN}
%  \label{tab:case_study}
%  \end{table*}

%  \begin{table*}[width=0.9\linewidth]
%  \centering
%  \begin{tabular}{llllllll}
% \hline
%  User & Age & Birthday & Gender & Marital & Province & City & description \\ \hline
%  U1 & 24 & 1994-5-20 & 2 & 0 & \begin{tabular}[c]{@{}l@{}}其他\\ Other\end{tabular} & NULL & \begin{tabular}[c]{@{}l@{}}只爱朱一龙\\ I just love Zhu Yilong\end{tabular} \\ \hline
%  U2 & 23 & 1995-4-28 & 2 & 1 & \begin{tabular}[c]{@{}l@{}}黑龙江\\ Heilongjiang\end{tabular} & NULL & \begin{tabular}[c]{@{}l@{}}努力成为更好的自己\\ To make myself become better\end{tabular} \\ \hline
%  U3 & 20 & 1998-3-06 & 2 & 0 & \begin{tabular}[c]{@{}l@{}}浙江\\ Zhejiang\end{tabular} & \begin{tabular}[c]{@{}l@{}}嘉兴\\ Jiaxing\end{tabular} & NULL \\ \hline
%  \end{tabular}
% \caption{Partial users Profile for case study sample}
%  \label{tab:user_prof}
% \end{table*}

\end{CJK}